\documentclass[sigconf]{acmart}
\AtBeginDocument{%
  }

\setcopyright{acmlicensed}
\copyrightyear{2018}
\acmYear{2018}
\acmDOI{XXXXXXX.XXXXXXX}
\acmConference[Conference acronym 'XX]{Make sure to enter the correct
  conference title from your rights confirmation email}{June 03--05,
  2018}{Woodstock, NY}
\acmISBN{978-1-4503-XXXX-X/2018/06}

\usepackage{hyperref}
\usepackage{url}

\usepackage{xr}
\usepackage{times}
\usepackage{latexsym}
\usepackage[T1]{fontenc}
\usepackage[utf8]{inputenc}
\usepackage{microtype}
\usepackage{graphicx}
\usepackage{mathtools}
\usepackage{dsfont}
\usepackage{algorithm}
\usepackage{algpseudocode}
\usepackage[capitalise]{cleveref}
\usepackage{multirow}
\usepackage{booktabs}
\usepackage{tabularx}
\usepackage{mathtools}

\begin{document}

\title{GMLM: Bridging Graph Neural Networks and Language Models for Heterophilic Node Classification}

\author{Aarush Sinha}
\email{aarush.sinha@gmail.com}
\affiliation{%
  \institution{Department of Computer Science ,University of Copenhagen}
  \city{Copenhagen}
  \country{Denmark}
}

\renewcommand{\shortauthors}{Sinha et al.}

\begin{abstract}
    Integrating Pre-trained Language Models (PLMs) with Graph Neural Networks (GNNs) remains a central challenge in text-rich heterophilic graph learning. We propose a novel integration framework that enables effective fusion between powerful pre-trained text encoders and Relational Graph Convolutional networks (R-GCNs). Our method enhances the alignment of textual and structural representations through a bi-directional fusion mechanism and contrastive node-level optimization. To evaluate the approach, we train two variants using different PLMs: \texttt{Snowflake-Embed} (state-of-the-art) and \texttt{GTE-base} each paired with an R-GCN backbone. Experiments on five heterophilic benchmarks demonstrate that our integration method achieves state-of-the-art results on four datasets, surpassing existing GNN and large language model–based approaches. Notably, \textit{Snowflake-Embed + R-GCN} improves accuracy on the Texas dataset by over 8\% and on Wisconsin by nearly 5\%. These results highlight the effectiveness of our fusion strategy for advancing text-rich graph representation learning.

\end{abstract}




\received{20 February 2007}
\received[revised]{12 March 2009}
\received[accepted]{5 June 2009}

\maketitle

\section{Introduction}
Graph Neural Networks (GNNs) \cite{4700287, zhou2021graphneuralnetworksreview} have demonstrated remarkable success in node classification, link prediction, and graph-level tasks. However, many such architectures focus primarily on structural information and local neighborhood features, potentially overlooking rich semantic content embedded within node attributes, especially textual attributes. While recent advances in natural language processing, particularly Pre-trained Language Models (PLMs) like BERT \cite{devlin-etal-2019-bert}, have revolutionized our ability to capture contextual information from text, their effective and scalable integration with graph learning remains an active area of research. The challenge is particularly pronounced in heterophilic graphs, where connected nodes may exhibit dissimilar features and labels, requiring a nuanced understanding of both structure and semantics.

While architectures like Graph Attention Networks (GATs) \cite{veličković2018graphattentionnetworks} offer learnable attention mechanisms for neighborhood aggregation, and Relational Graph Convolutional Networks (RGCNs) \cite{schlichtkrull2017modelingrelationaldatagraph} can handle typed edges, they do not inherently perform deep semantic reasoning on node-associated texts without explicit integration with PLMs. Concurrently, masked modeling techniques, central to the success of PLMs, have proven highly effective for learning contextual representations. This prompts the question: How can we develop a deeply integrated framework that combines the strengths of GNNs and PLM-style masked modeling to enhance representation learning on text-rich graphs ?

Traditional node feature masking in GNNs \cite{mishra2021nodemaskingmakinggraph} often involves random feature perturbation or complete removal, which can disrupt learning or fail to provide sufficiently informative training signals. Furthermore, applying PLMs to all node texts in large graphs is often computationally prohibitive.

In this paper, we present a novel framework that deeply integrates GNNs and PLMs for improved node representation learning on text-rich graphs, with a focus on heterophilic settings. Our approach introduces three key innovations:
\begin{enumerate}
    \item[(1)] A \textbf{dynamic active node selection strategy} that stochastically identifies a subset of nodes during each training iteration. By processing only the texts of these active nodes with the PLM, we drastically reduce the computational overhead and memory footprint, making the approach scalable.
    \item[(2)] A \textbf{GNN-specific contrastive pre-training stage} that pre-trains a deep, multi-scale GNN (an RGCN in our implementation) to learn robust structural representations. This stage employs a novel soft masking mechanism that interpolates original node features with a learnable, graph-specific [MASK] token embedding, providing a more stable and informative training signal than hard masking.
    \item[(3)] A \textbf{bi-directional cross-attention fusion module} that moves beyond simple concatenation. It allows the multi-scale graph embeddings and the deep semantic PLM embeddings to mutually inform and refine one another, creating a rich, contextually-aware joint representation before the final classification.
\end{enumerate}

\section{Related Works}

Graph Neural Networks (GNNs) \cite{zhou2021graphneuralnetworksreview} provide a powerful framework for learning from graph-structured data. Building upon early spectral and spatial convolutional methods, Graph Attention Networks (GATs) \cite{veličković2018graphattentionnetworks} integrated self-attention \cite{10.5555/3295222.3295349}. This allows adaptive weighting of neighbors during message passing, enhancing model expressivity and suitability for irregular graph structures, as detailed in recent surveys.

Efforts to adapt the Transformer architecture, initially designed for sequences, to graph data aim to capture long-range dependencies via global self-attention, potentially overcoming limitations of traditional message passing \cite{ying2021transformersreallyperformbad}, \cite{yun2020graphtransformernetworks}. These Graph Transformers, reviewed systematically \cite{shehzad2024graphtransformerssurvey}, can incorporate positional information, handle diverse graph scales, achieve competitive results, and utilize language model pretraining benefits on text-rich graphs.

\textbf{Masked Pretraining Strategies}, popularized by language models \cite{devlin-etal-2019-bert} that learn contextual representations via masked token prediction, have been adapted for graphs. The Masked Graph Autoencoder (MaskGAE) \cite{mgae} framework, for instance, masks graph edges and trains the model to reconstruct them, thereby learning robust representations for downstream tasks like link prediction and node classification.

\textbf{Integration of Language Models and Graph Structures} seeks to combine textual and structural information. Key directions include: fusing pretrained language models (LMs) with GNNs for joint reasoning over text and structure \cite{plenz-frank-2024-graph}; injecting structural information directly into LMs using graph-guided attention mechanisms, potentially removing the need for separate GNN modules \cite{yuan-farber-2024-grasame}; interpreting Transformers as GNNs operating on complete graphs, allowing explicit modeling of edge information \cite{henderson-etal-2023-transformers}; and jointly training large LMs and GNNs to utilize both contextual understanding and structural learning capabilities \cite{ioannidis2022efficienteffectivetraininglanguage}. 

\begin{figure*}[!ht]
    \centering
    \includegraphics[width=\textwidth]{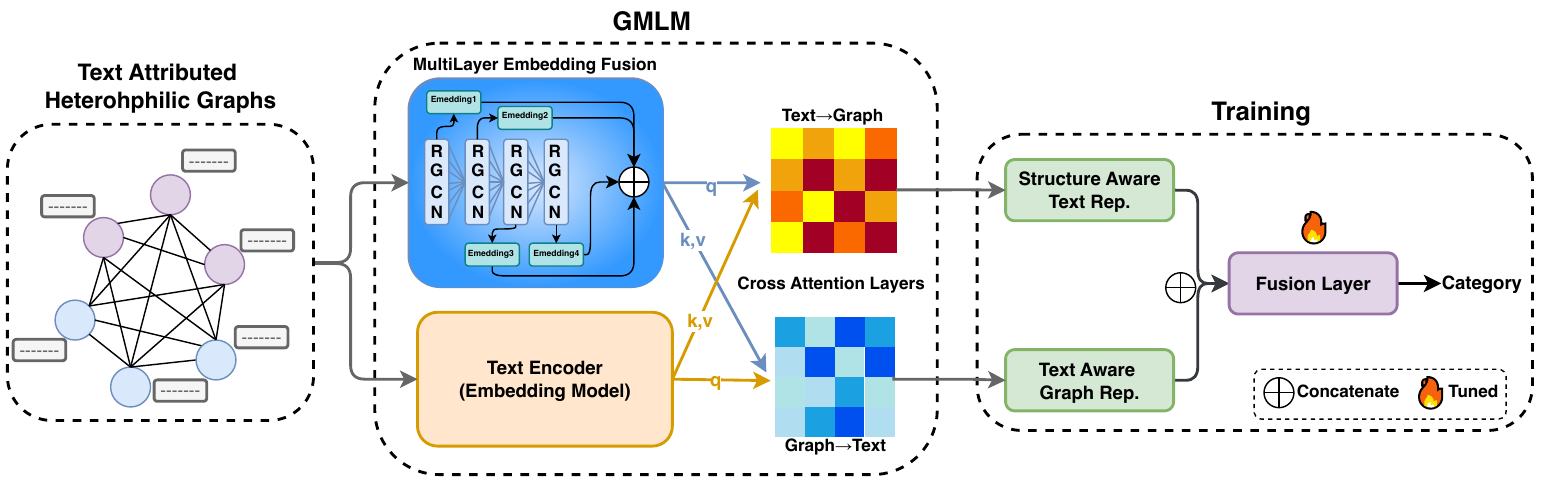}
    \caption{The text related to each node is fed into the text-encoder and the node features by the RGCN layers that use multi-layer embeddings fused together. The Cross Attention layers between Graph and Text are then concatenated by a fusion layer which is finally trained and evaluated.}
    \label{fig:method}
\end{figure*}

\section{Preliminaries}
\label{sec:preliminaries}

In this section, we introduce the fundamental concepts and notations used throughout this paper, including graph representation, Graph Neural Networks (specifically RGCNs), Pre-trained Language Models, and the concept of heterophily in graphs.

\subsection{Relational Graph Convolutional Networks (RGCNs)}
Relational Graph Convolutional Networks (RGCNs) \cite{schlichtkrull2017modelingrelationaldatagraph} extend GCNs to handle multi-relational graphs (heterogeneous graphs with different edge types). The message passing for a node $v_i$ in an RGCN layer is typically defined as:
\begin{equation}
    \mathbf{h}_i^{(l+1)} = \sigma \left( \sum_{r \in \mathcal{R}} \sum_{v_j \in \mathcal{N}_r(v_i)} \frac{1}{c_{i,r}} \mathbf{W}_r^{(l)} \mathbf{h}_j^{(l)} + \mathbf{W}_0^{(l)} \mathbf{h}_i^{(l)} \right)
    \label{eq:rgcn}
\end{equation}
where $\mathcal{N}_r(v_i)$ is the set of neighbors of node $v_i$ under relation $r \in \mathcal{R}$, $\mathbf{W}_r^{(l)}$ is a relation-specific learnable weight matrix for relation $r$ at layer $l$, $\mathbf{W}_0^{(l)}$ is a weight matrix for self-loops, $c_{i,r}$ is a normalization constant ( $|\mathcal{N}_r(v_i)|$), and $\sigma$ is an activation function ( ReLU). To handle a large number of relations, RGCNs often use basis decomposition or block-diagonal decomposition for $\mathbf{W}_r^{(l)}$.

\subsection{Pre-trained Language Models (PLMs)}
Pre-trained Language Models (PLMs), such as BERT \cite{devlin-etal-2019-bert} and its variants (DistilBERT \cite{sanh2020distilbertdistilledversionbert}), have achieved state-of-the-art results on various natural language processing (NLP) tasks. These models are typically based on the Transformer architecture \cite{vaswani2023attentionneed} and are pre-trained on massive text corpora using self-supervised objectives like Masked Language Modeling (MLM) and Next Sentence Prediction (NSP).
Given an input text sequence ( a node's textual description $t_i$), a PLM first tokenizes it into a sequence of sub-word tokens $[w_1, w_2, \dots, w_S]$. These tokens are then converted into input embeddings (sum of token, position, and segment embeddings) and fed into multiple Transformer layers. The output of the PLM is a sequence of contextualized embeddings for each token. For sentence-level or document-level tasks, often the embedding of a special [CLS] token ( $\mathbf{h}_{[CLS]}$) or an aggregation ( mean pooling) of token embeddings is used as the representation of the entire input text.
In our work, we utilize a PLM to obtain semantic embeddings $\mathbf{h}_{PLM}(t_i)$ from node texts $t_i$.

\subsection{Node Feature Masking in GNNs}
Masking is a common technique in self-supervised learning. In the context of GNNs, node feature masking aims to train the model to reconstruct or predict properties of masked nodes or their features, thereby learning robust representations. A simple approach might involve replacing a fraction of node features $\mathbf{x}_i$ with a zero vector or a special [MASK] token embedding before feeding them into the GNN \cite{gptgnn, NEURIPS2020_3fe23034}. Our GNN contrastive pretraining phase builds upon this idea but introduces a soft masking mechanism with a learnable graph-specific [MASK] token.

\subsection{Heterophily in Graphs}
Graph homophily refers to the principle that nodes tend to connect to other nodes that are similar to themselves ( in terms of features or labels). Conversely, graph heterophily (or disassortativity) describes the scenario where connected nodes are often dissimilar \cite{hetero}. Many real-world graphs exhibit heterophily, such as protein-protein interaction networks or certain social networks. Standard GNNs, which assume homophily by smoothing features over neighborhoods, may perform poorly on heterophilic graphs. Designing GNN architectures that can effectively handle heterophily is an ongoing research challenge, often requiring models to capture more complex relational patterns beyond simple neighborhood similarity. Our work aims to contribute to this by integrating rich textual semantics which can provide crucial context in such settings.

\subsection{Contrastive Learning}
Contrastive learning is a self-supervised learning paradigm that aims to learn representations by maximizing agreement between differently augmented "views" of the same data sample (positive pairs) while minimizing agreement between views of different samples (negative pairs). A common contrastive loss function is the NT-Xent loss \cite{NIPS2016_6b180037}:
\begin{equation}
    \mathcal{L}_{NT\text{-}Xent} = - \sum_{i=1}^{B} \log \frac{\exp(\text{sim}(\mathbf{z}_i, \mathbf{z}_j) / \tau)}{\sum_{k=1}^{2B} \mathbb{I}_{[k \neq i]} \exp(\text{sim}(\mathbf{z}_i, \mathbf{z}_k) / \tau)}
\label{eq:ntxent}
\end{equation}
where $(\mathbf{z}_i, \mathbf{z}_j)$ is a positive pair of augmented representations from the same original sample within a batch of size $B$, $\text{sim}(\cdot, \cdot)$ is a similarity function ( cosine similarity), $\tau$ is a temperature hyperparameter, and the sum in the denominator is over all $2B$ augmented samples in the batch, excluding $\mathbf{z}_i$ itself. Our GNN pretraining phase utilizes this principle.

\section{Methodology}
\label{sec:methodology}

We introduce a novel architecture designed to effectively integrate deep graph structural information with rich textual semantics from nodes for robust node representation learning. The architecture is characterized by two main processing pathways for graph and text data, a multi-scale GNN with residual connections, a sophisticated bi-directional cross-attention fusion module, and specialized input perturbation mechanisms to enhance robustness and training stability. We provide an overview of our proposed architecture in Figure \ref{fig:method}. All experiments were done on a single RTX8000(45GB) using PyTorch \cite{pytorch} and PyG\cite{Feyetal2025}\cite{FeyLenssen2019}.

\subsection{GNN Input Feature Perturbation with Soft Masking}
\label{ssec:soft_masking}
To enhance the robustness of the GNN and to provide a meaningful training signal, we employ a soft masking strategy for the input node features. This is applied during both a contrastive pre-training phase and the final end-to-end fine-tuning. Let $\mathbf{X} \in \mathbb{R}^{N \times d_x}$ be the matrix of initial node features for $N$ nodes. Given a binary perturbation mask $\mathbf{m}_{pert} \in \{0,1\}^N$, the soft masking operation produces perturbed features $\mathbf{X}'$ as follows:
\begin{equation}
    \mathbf{x}'_i =
    \begin{cases}
        (1-\beta) \cdot \mathbf{x}_i + \beta \cdot \mathbf{e}_{mask}^{gnn} & \text{if } m_{pert,i} = 1 \\
        \mathbf{x}_i & \text{if } m_{pert,i} = 0
    \end{cases}
    \label{eq:soft_masking}
\end{equation}
where $\mathbf{x}_i$ is the original feature vector for node $v_i$, $\mathbf{e}_{mask}^{gnn} \in \mathbb{R}^{d_x}$ is a learnable \textbf{GNN-specific mask token embedding}, and $\beta \in [0,1]$ is a hyperparameter controlling the interpolation weight. This GNN-specific mask token $\mathbf{e}_{mask}^{gnn}$ is initialized using Xavier uniform initialization and learned alongside other model parameters.

\subsection{Multi-Scale Relational GNN Branch}
\label{ssec:gnn_branch}
The GNN branch is designed as a deep, multi-scale architecture to capture structural information across different neighborhood depths. It utilizes a stack of four Relational Graph Convolutional Network (RGCN) layers. The input to the first RGCN layer is the perturbed node feature matrix $\mathbf{X}'$ from Equation~\ref{eq:soft_masking}.

Each of the four GNN blocks consists of an RGCNConv layer, followed by GraphNorm, a GELU activation function, and Dropout. To facilitate the training of this deep architecture and combine features from different network depths, we incorporate residual connections with linear projections. The hidden representation $\mathbf{H}^{(l)}$ at layer $l$ is computed as:
\begin{align}
    \mathbf{Z}^{(l)} &= \text{GELU}(\text{GraphNorm}(\text{RGCN}^{(l)}(\mathbf{H}^{(l-1)}, \mathcal{E}))) \\
    \mathbf{H}^{(l)} &= \text{Dropout}(\mathbf{Z}^{(l)}) + \text{Proj}^{(l-1)}(\mathbf{H}^{(l-2)})
\end{align}
where $\mathbf{H}^{(0)} = \mathbf{X}'$, $\mathcal{E}$ represents the graph edges and their types, and $\text{Proj}^{(l-1)}$ is a linear projection. This structure allows the model to maintain and propagate information from earlier layers. We employ gradient checkpointing\footnote{Specifically, torch.utils.checkpoint.checkpoint is used for each GNN block to trade a small amount of re-computation for significant memory savings, enabling the use of a deeper GNN.} on each GNN block to manage memory consumption during backpropagation.

Instead of using only the output of the final GNN layer, we adopt a \textbf{multi-scale fusion} strategy to create a more comprehensive graph representation. The embeddings from all four GNN layers, $\{\mathbf{H}^{(1)}, \mathbf{H}^{(2)}, \mathbf{H}^{(3)}, \mathbf{H}^{(4)}\}$, are collected. These are then combined using a learnable weighted sum after being projected to the PLM's hidden dimension, $d_{PLM}$:
\begin{equation}
    \mathbf{h}_{G,i} = \text{LayerNorm}\left(\sum_{l=1}^{4} w_l \cdot \text{Proj}^{(l)}(\mathbf{h}_{i}^{(l)})\right)
    \label{eq:multiscale_fusion}
\end{equation}
where $\mathbf{h}_{i}^{(l)}$ is the embedding for node $v_i$ from layer $l$, $\text{Proj}^{(l)}$ is a linear projection to dimension $d_{PLM}$, and $w_l$ are learnable weights derived from a softmax-normalized parameter vector. The final graph embedding for node $v_i$ is $\mathbf{h}_{G,i} \in \mathbb{R}^{d_{PLM}}$.

\subsection{Pre-trained Language Model (PLM) Branch}
\label{ssec:plm_branch}
The PLM branch processes the textual descriptions associated with nodes to capture rich semantic information. To manage computational costs, especially on large graphs, we process texts in micro-batches. During each training iteration, an \textbf{active node mask} $\mathbf{m}_{active} \in \{0,1\}^N$ identifies a subset of nodes $\mathcal{V}_{active}$ whose texts will be processed.

For each node $v_i \in \mathcal{V}_{active}$, its text $t_i$ is tokenized and fed into a pre-trained language model, such as \textbf{GTE-Base (110M)\cite{li2023generaltextembeddingsmultistage} and Snowflake-Embed(305M)\cite{yu2024arcticembed20multilingualretrieval}}. To derive a fixed-size representation from the PLM's variable-length output, we perform \textbf{mean pooling} over the last hidden state, weighted by the attention mask to exclude padding tokens. The PLM embedding for an active node $v_i$ is:
\begin{equation}
    \mathbf{h}_{PLM,i} = \frac{\sum_{j=1}^{L} \mathbf{o}_{j} \cdot a_{j}}{\sum_{j=1}^{L} a_{j}}
    \label{eq:plm_embedding}
\end{equation}
where $\mathbf{O} = \text{PLM-Encoder}(t_i)$ is the last hidden state sequence of length $L$, $\mathbf{o}_j$ is the embedding of the $j$-th token, and $a_j$ is the corresponding value from the attention mask (1 for real tokens, 0 for padding). For inactive nodes $v_k \notin \mathcal{V}_{active}$, their PLM embeddings are set to zero vectors, resulting in a complete PLM embedding matrix $\mathbf{H}_{PLM} \in \mathbb{R}^{N \times d_{PLM}}$.

\subsection{Bi-Directional Attention Fusion and Classification}
\label{ssec:fusion_classification}
To achieve a deep and meaningful integration of graph and text modalities, we move beyond simple concatenation and employ a \textbf{bi-directional cross-attention} mechanism. This allows the graph and text representations to mutually inform and refine one another. The full node embedding matrices, $\mathbf{H}_{G}$ (from Section~\ref{ssec:gnn_branch}) and $\mathbf{H}_{PLM}$ (from Section~\ref{ssec:plm_branch}), serve as inputs.

The cross-attention mechanism computes two sets of attended features:
\begin{enumerate}
    \item \textbf{Graph-to-Text Attention}: The PLM embeddings serve as queries to attend to the GNN embeddings, producing text-aware graph features, $\mathbf{H}_{G \to T}$.
    \item \textbf{Text-to-Graph Attention}: The GNN embeddings serve as queries to attend to the PLM embeddings, producing structure-aware text features, $\mathbf{H}_{T \to G}$.
\end{enumerate}
The standard scaled dot-product attention is used, where for a query matrix $\mathbf{Q}$, key matrix $\mathbf{K}$, and value matrix $\mathbf{V}$:
\begin{equation}
    \text{Attention}(\mathbf{Q}, \mathbf{K}, \mathbf{V}) = \text{softmax}\left(\frac{\mathbf{Q}\mathbf{K}^T}{\sqrt{d_k}}\right)\mathbf{V}
\end{equation}
The two attended representations are then concatenated for each node:
\begin{equation}
    \mathbf{h}_{concat,i} = [\mathbf{h}_{G \to T, i} \,;\, \mathbf{h}_{T \to G, i}] \in \mathbb{R}^{2 \cdot d_{PLM}}
    \label{eq:concat}
\end{equation}
This concatenated vector is passed through a fusion network, which is a feed-forward network consisting of a linear layer, LayerNorm, a GELU activation, and Dropout, to produce the final integrated representation:
\begin{equation}
    \mathbf{h}_{fused,i} = \text{FusionNet}(\mathbf{h}_{concat,i}) \in \mathbb{R}^{d_{fused}}
    \label{eq:fusion_mlp}
\end{equation}
Finally, this fused representation $\mathbf{h}_{fused,i}$ is fed into a classification head, another MLP (Linear -> GELU -> Dropout -> Linear), to predict the node label probabilities:
\begin{equation}
    \hat{\mathbf{y}}_i = \text{Softmax}(\text{MLP}_{classifier}(\mathbf{h}_{fused,i}))
    \label{eq:classifier}
\end{equation}
where $\hat{\mathbf{y}}_i \in \mathbb{R}^{|\mathcal{Y}|}$ contains the predicted probabilities for each class in the label set $\mathcal{Y}$.

\section{Training Procedure}
\label{sec:training_procedure}

The model training proceeds in two distinct stages: (1) GNN Contrastive Pre-training, focused on learning robust structural representations and initializing the GNN-specific mask token, followed by (2) End-to-End Fine-tuning, where the entire model is trained for the downstream node classification task. This two-stage approach allows the GNN to develop a strong structural foundation before being integrated with the powerful but computationally intensive language model. All training was done on a single Nvidia Quadro RTX 8000 (48GB).

\subsection{Stage 1: GNN Contrastive Pre-training}
\label{sec:gnn_pretraining_stage}
The primary goal of this stage is to pre-train the multi-scale GNN branch (Section~\ref{ssec:gnn_branch}) and the learnable GNN-specific mask token embedding $\mathbf{e}_{mask}^{gnn}$ (Section~\ref{ssec:soft_masking}) to capture meaningful structural information. This is achieved through a contrastive learning objective. During this stage, the weights of the PLM encoder, fusion network, and final classifier are frozen.

\subsubsection{Data Augmentation via Soft Masking}
For each training iteration, we generate two augmented "views" of the graph. Two distinct perturbation masks, $\mathbf{m}_{pert,1}$ and $\mathbf{m}_{pert,2}$, are created by selecting a proportion $p_{gnn\_mask}$ (randomly sampled from $\mathcal{U}(0.2, 0.4)$ each iteration) of all nodes in the graph. This selection is weighted by node degree to prioritize more connected nodes. Using these masks and the soft masking mechanism (Equation~\ref{eq:soft_masking}) with an interpolation weight $\beta=0.7$, we obtain two sets of perturbed node features, $\mathbf{X}'_1$ and $\mathbf{X}'_2$.

\subsubsection{Contrastive Objective}
Both $\mathbf{X}'_1$ and $\mathbf{X}'_2$ are passed through the \emph{same shared-weight GNN encoder} to produce two sets of multi-scale node embeddings, $\mathbf{H}_{G,1}$ and $\mathbf{H}_{G,2}$. We then apply the NT-Xent contrastive loss with a temperature $\tau_{pretrain}=0.1$. The pre-training loss is:
\begin{equation}
    \mathcal{L}_{pretrain} = \mathcal{L}_{NT-Xent}(\mathbf{H}_{G,1}, \mathbf{H}_{G,2}, \tau_{pretrain})
    \label{eq:pretrain_loss}
\end{equation}
This pre-training is conducted for \textbf{30 epochs}. Only the parameters of the GNN branch (RGCN layers, graph normalization, residual projections, multi-scale fusion) and the mask token $\mathbf{e}_{mask}^{gnn}$ are updated. We use an AdamW optimizer with a learning rate of $1 \times 10^{-4}$ for all trainable parameters in this stage and a weight decay of $0.05$. A CosineAnnealingWarmRestarts learning rate scheduler is employed to adjust the learning rate throughout pre-training.

\subsection{Stage 2: End-to-End Fine-tuning for Node Classification}
\label{sec:end_to_end_finetuning}
After GNN pre-training, the entire model, with the GNN branch initialized from Stage 1, is fine-tuned for node classification. The model is designed to leverage raw textual information associated with nodes, which is particularly valuable for navigating the complex relationships in heterophilic graphs.

\subsubsection{Active Node Selection and GNN Perturbation}
For each training iteration, an active node mask $\mathbf{m}_{active}$ is generated. This mask first considers nodes in the training split ($\mathcal{V}_{train}$). From this set, a random proportion $p_{active\_nodes}$ (sampled from $\mathcal{U}(0.3, 0.8)$) is selected based on node degree. This active mask serves two purposes:
\begin{enumerate}
    \item \textbf{PLM Processing:} Only the texts $\{t_i : m_{active,i}=1\}$ are processed by the PLM to manage computational load.
    \item \textbf{GNN Input Perturbation:} The same mask $\mathbf{m}_{active}$ is used to apply soft masking (Equation~\ref{eq:soft_masking}) to the GNN's input features, continuing to regularize the GNN during fine-tuning.
\end{enumerate}
The final classification loss is also computed exclusively over these active nodes.

\subsubsection{Supervised Classification Loss}
The model processes the inputs to produce predicted label probabilities $\hat{\mathbf{y}}_i$ (Equation~\ref{eq:classifier}). The supervised classification loss $\mathcal{L}_{classify}$ is the cross-entropy loss with label smoothing (smoothing factor of 0.2) applied to the true labels.

\subsubsection{Optimization and Hyperparameters}
The total loss $\mathcal{L}_{classify}$ is backpropagated through the entire model. All trainable parameters are updated using the AdamW optimizer \cite{loshchilov2018decoupled} with differential learning rates:
\begin{itemize}
    \item GNN components (RGCN layers, projections, etc.): $lr_{graph} = 1 \times 10^{-4}$.
    \item PLM ({GTE-Base}) parameters: $lr_{bert} = 1 \times 10^{-5}$.
    \item Other components (attention, fusion, classifier, mask token): $lr_{other} = 1 \times 10^{-4}$.
\end{itemize}
A weight decay of $0.05$ is applied to the GNN and "other" parameter groups, while the PLM uses a standard weight decay of $0.01$. The model is trained for a maximum of \textbf{500 epochs}. A linear learning rate scheduler with a warm-up period corresponding to the first 10\% of training steps is used. Additionally, gradient clipping with a maximum norm of 1.0 is applied to prevent exploding gradients. We employ an early stopping strategy with a patience of \textbf{30 epochs}: training halts if the F1 score on the validation set does not improve. The model state with the best validation F1 score is used for the final evaluation.

The key hyperparameters are summarized below:
\begin{itemize}
    \item \textbf{$p_{gnn\_mask}$ (Pre-training Perturbation Ratio):} Sampled from $\mathcal{U}(0.2, 0.4)$ in each pre-training epoch.
    \item \textbf{$p_{active\_nodes}$ (Fine-tuning Active Node Ratio):} Sampled from $\mathcal{U}(0.3, 0.8)$ in each fine-tuning epoch.
    \item \textbf{$\beta$ (Soft Masking Weight):} Set to $0.7$ for both stages.
    \item \textbf{$\tau_{pretrain}$ (Pre-training Contrastive Temperature):} Set to $0.1$.
\end{itemize}

\begin{table*}[!ht]
\centering
\begin{tabular}{lccccc}
\toprule
\textbf{Methods} & \textbf{Cornell} & \textbf{Texas} & \textbf{Wisconsin} & \textbf{Actor} & \textbf{Amazon} \\
\midrule
\multicolumn{6}{c}{\textit{Classic GNNs}} \\
\midrule
GCN \cite{kipf2017semisupervisedclassificationgraphconvolutional} & 52.86$\pm$1.8 & 43.64$\pm$3.3 & 41.40$\pm$1.8 & 66.70$\pm$1.3 & 39.33$\pm$1.0 \\
GraphSAGE\cite{hamilton2017inductive} & 75.71$\pm$1.8 & 81.82$\pm$2.5 & 80.35$\pm$1.3 & 70.37$\pm$0.1 & 46.63$\pm$0.1 \\
GAT\cite{veličković2018graphattentionnetworks} & 54.28$\pm$5.1 & 51.36$\pm$2.3 & 50.53$\pm$1.7 & 63.74$\pm$6.7 & 35.12$\pm$6.4 \\
\midrule
\multicolumn{6}{c}{\textit{Heterophily-specific GNNs}} \\
\midrule
H2GCN\cite{zhu2020beyond} & 69.76$\pm$3.0 & 79.09$\pm$3.5 & 80.18$\pm$1.9 & 70.73$\pm$0.9 & 47.09$\pm$0.3 \\
FAGCN\cite{bo2021lowfrequencyinformationgraphconvolutional} & 76.43$\pm$3.1 & 84.55$\pm$4.8 & 83.16$\pm$1.4 & 75.58$\pm$0.5 & 49.83$\pm$0.6 \\
JacobiConv\cite{wang2022powerfulspectralgraphneural} & 73.57$\pm$4.3 & 81.80$\pm$4.1 & 76.31$\pm$11.3 & 73.81$\pm$0.3 & 49.43$\pm$0.5 \\
GBK-GNN\cite{du2022gbkgnngatedbikernelgraph} & 66.19$\pm$2.8 & 80.00$\pm$3.0 & 72.98$\pm$3.3 & 72.49$\pm$1.0 & 44.90$\pm$0.3 \\
OGNN\cite{Wang_2022} & 71.91$\pm$1.8 & 85.00$\pm$2.3 & 79.30$\pm$2.1 & 72.08$\pm$2.4 & 47.79$\pm$1.6 \\
SEGSL\cite{DBLP:conf/www/ZouPHYLWLY23} & 66.67$\pm$4.1 & 85.00$\pm$2.0 & 79.30$\pm$1.8 & 72.73$\pm$0.8 & 47.38$\pm$0.2 \\
DisamGCL\cite{disam} & 50.48$\pm$2.0 & 65.00$\pm$1.2 & 57.89$\pm$0.0 & 67.78$\pm$0.3 & 43.90$\pm$0.4 \\
\midrule
\multicolumn{6}{c}{\textit{LLM4HeG \cite{wu2025exploringpotentiallargelanguage} (fine-tuned LLM/SLMs and distilled SLMs)}} \\
\midrule
Vicuna 7B  & \underline{77.62$\pm$2.9} & \underline{89.09$\pm$3.3} & 86.14$\pm$2.1 & \textbf{76.82$\pm$0.5} & \underline{51.53$\pm$0.4} \\
Bloom 560M  & 75.48$\pm$2.1 & 80.00$\pm$4.0 & {86.49}$\pm$1.9 & {76.16}$\pm$0.6 & 51.52$\pm$0.5 \\
Bloom 1B & 75.71$\pm$1.4 & 83.86$\pm$2.8 & 83.86$\pm$1.7 & 74.99$\pm$0.5 & {52.33}$\pm$0.6 \\
7B-to-560M & 75.00$\pm$4.0 & {88.18}$\pm$2.2 & {87.19}$\pm$2.5 & 75.78$\pm$0.2 & 51.51$\pm$0.4 \\
7B-to-1B & {77.38}$\pm$2.7 & {88.18}$\pm$4.0 & 86.14$\pm$1.5 & 75.37$\pm$0.9 & {51.58}$\pm$0.4 \\
\midrule
\multicolumn{6}{c}{\textit{GMLM}} \\
\midrule
GTE-Base & {74.25}$\pm$2.5 & 85.90$\pm$4.9 & \underline{89.81$\pm$2.5} & 73.89$\pm$2.1 & 50.64$\pm$0.7 \\
Snowflake-Embed & \textbf{79.16$\pm$4.8} & \textbf{97.18$\pm$2.9} & \textbf{92.04$\pm$1.6} & \underline{76.81$\pm$0.9} & \textbf{54.63$\pm$0.4} \\
\bottomrule
\end{tabular}
\caption{Accuracy for node classification of different methods. \textbf{Bold} denotes the best score while \textit{underline} denotes the second best.}
\label{tab:results_comparison}
\end{table*}

\section{Results}

We evaluate GMLM on five widely-used heterophilic benchmark datasets. Following standard protocols, we randomly split the nodes into train, validation, and test sets with a proportion of 48\%/32\%/20\% for the Cornell, Texas, Wisconsin, and Actor datasets, and 50\%/25\%/25\% for Amazon following \cite{platonov2024criticallookevaluationgnns}. All reported results in Table~\ref{tab:results_comparison} are the mean accuracy (\%) and standard deviation averaged over 10 independent runs.

As detailed in the table, our GMLM framework achieves state-of-the-art performance on the majority of benchmarks. The GMLM-Snowflake-embed variant establishes new records on four of the five datasets: Cornell, Texas, Wisconsin, and Amazon. The performance gains are particularly striking on the Texas dataset, where our model achieves 97.18\% accuracy, surpassing the strongest LLM-based baseline by over 8 absolute percentage points. Similarly, it improves the state-of-the-art on Wisconsin by nearly 5 points to 92.04\%.

Crucially, GMLM consistently outperforms both specialized heterophily-GNNs and even the much larger Vicuna 7B model. This demonstrates that our architectural design, which deeply integrates a pre-trained GNN with a PLM via bi-directional cross-attention, is more effective and capital-efficient than simply applying a larger, general-purpose model. On the Actor dataset, our result is highly competitive, nearly matching the 7B model's performance with a significantly smaller PLM backbone.

A comparison of our model variants reveals that performance scales with the representational power of the PLM, as Snowflake-embed (305M) consistently outperforms GTE-Base (110M). However, even the smaller GMLM-GTE-Base is highly competitive, outperforming the majority of all baselines and securing the second-best result on Wisconsin. This confirms the robustness and effectiveness of the core GMLM framework itself. We provide a full qualitative comparison of the PLM backbones and the effect of different training stages via UMAP visualizations on Cornell, Texas and Wisconsin in Section \ref{sec:ets}.

\section{Effect of Training Stages}
\label{sec:ets}

\subsection{Cornell}

We study the effect of different stages of training our GMLM framework with the different PLM backbones. Figure \ref{fig:cg} and Figure \ref{fig:cs} show the embedding visualizations of the node embeddings on the Cornell dataset, projected into a 2D space using UMAP. The plots illustrate the state of embeddings for various node types (students, courses, etc.) at three key stages: initial raw features, after contrastive pre-training, and after full fine-tuning. Figure \ref{fig:cg} presents the results using the GTE-base model as the text encoder. While it shows a noticeable improvement in cluster formation from the initial state to the fine-tuned stage, the final clusters remain somewhat diffuse and exhibit overlap. In contrast, Figure \ref{fig:cs}, which uses the larger Snowflake embedding model, demonstrates markedly superior performance. Following the full fine-tuning process, the node clusters are significantly more compact, distinct, and well-separated. This visual evidence indicates that the Snowflake model generates more discriminative feature representations and strongly supports the conclusion that it serves as a more powerful text encoder for this particular dataset and task.

\begin{figure*}[!ht]
    \centering
    \includegraphics[width=\linewidth]{images/Cornell_embedding_ablation_gte.png}
    \caption{UMAP visualization of node embeddings using the GTE-base model (PLM backbone). The three panels show the embeddings for raw features, after contrastive pre-training, and after full fine-tuning.}
    \label{fig:cg}
\end{figure*}

\begin{figure*}[!ht]
    \centering
    \includegraphics[width=\linewidth]{images/Cornell_embedding_ablation_snow.png}
    \caption{UMAP visualization of node embeddings using the larger Snowflake embedding model (PLM backbone). These panels show a clear improvement in cluster separation after fine-tuning compared to the GTE-base model in Figure \ref{fig:cg}.}
    \label{fig:cs}
\end{figure*}

\subsection{Texas}
\label{ssec:texas_viz}

Similarly, we analyze the training effects on the Texas dataset. Figure \ref{fig:tg} and Figure \ref{fig:ts} display the UMAP visualizations of node embeddings, highlighting the progression from initial features through contrastive pre-training to full fine-tuning. In Figure \ref{fig:tg}, the GTE-base model, while improving cluster separation compared to the raw features, still results in embeddings where the classes are not sharply delineated after fine-tuning. Conversely, the larger Snowflake embedding model, shown in Figure \ref{fig:ts}, yields significantly better results. After the full fine-tuning stage, the node clusters are dense and clearly separated, demonstrating the superior capability of the Snowflake model to learn discriminative representations for the node types within the Texas dataset. This again underscores the impact of the PLM backbone's scale and architecture on the final performance of the GMLM framework.

\begin{figure*}[!ht]
    \centering
    \includegraphics[width=\textwidth]{images/Texas_embedding_ablation_gte.png}
    \caption{UMAP visualization of node embeddings on the Texas dataset using the GTE-base model. The plots show the progression from raw features to fine-tuned embeddings.}
    \label{fig:tg}
\end{figure*}

\begin{figure*}[!ht]
    \centering
    \includegraphics[width=\textwidth]{images/Texas_embedding_ablation_snow.png}
    \caption{UMAP visualization of node embeddings on the Texas dataset using the Snowflake embedding model. A clear improvement in cluster quality is visible compared to the GTE-base model shown in Figure \ref{fig:tg}.}
    \label{fig:ts}
\end{figure*}

\begin{figure*}[!ht]
    \centering
    \includegraphics[width=\textwidth]{images/Wisconsin_embedding_ablation_gte.png}
    \caption{UMAP visualization of node embeddings on the Wisconsin dataset using the GTE-base model. The panels show the progression from raw features to fine-tuned embeddings, resulting in moderately separated clusters.}
    \label{fig:wg}
\end{figure*}

\begin{figure*}[!ht]
    \centering
    \includegraphics[width=\textwidth]{images/Wisconsin_embedding_ablation_snow.png}
    \caption{UMAP visualization of node embeddings on the Wisconsin dataset using the Snowflake embedding model. A dramatic improvement in cluster density and separation is visible compared to the GTE-base model shown in Figure \ref{fig:wg}.}
    \label{fig:ws}
\end{figure*}

\subsection{Wisconsin}

Finally, we present the results for the Wisconsin dataset. The visualization of node embeddings in Figure \ref{fig:wg} and Figure \ref{fig:ws} follows the same three-stage training progression. For the GTE-base model (Figure \ref{fig:wg}), the fine-tuning process yields some structure, but the resulting clusters lack clear separation and show considerable overlap between different node types. In stark contrast, the Snowflake embedding model (Figure \ref{fig:ws}) demonstrates exceptional performance. The clusters corresponding to different node types are highly compact and distinctly separated after full fine-tuning. This comparison on the Wisconsin dataset further reinforces our finding that the larger PLM backbone is crucial for learning high-quality, discriminative node representations within our GMLM framework.

\section{Conclusion}
\label{sec:conclusion}

In this paper, we introduced the Graph Masked Language Model (GMLM), a novel framework that effectively integrates deep graph structural learning with the rich semantic understanding of Pre-trained Language Models. By employing a two-stage training procedure featuring a contrastive GNN pre-training phase and an end-to-end fine-tuning stage with dynamic active node selection, our model addresses key challenges of scalability and robust representation learning. The use of a sophisticated bi-directional cross-attention module for fusion proved critical, enabling GMLM to achieve new state-of-the-art results on four out of five widely-used heterophilic benchmarks. Our model significantly outperforms both specialized GNNs and much larger LLM-based approaches, underscoring that a carefully designed, deeply integrated architecture can be more powerful and capital-efficient than brute-force scaling for text-rich graph representation learning.

\section{Limitations and Future Work}
\label{sec:limitations}

Despite the strong performance of GMLM, we acknowledge the limitations that present opportunities for future research. The framework's effectiveness is intrinsically tied to the availability and quality of textual descriptions for nodes. In scenarios where such texts are sparse, noisy, or entirely absent, the PLM branch offers diminished value. Future work could explore robust fallbacks or methods for generating informative pseudo-texts to enhance performance on text-poor graphs.

\bibliographystyle{ACM-Reference-Format}
\bibliography{arxiv}


\end{document}